\documentclass[10pt,twocolumn,letterpaper]{article}
\usepackage[pagenumbers]{cvpr} 

\usepackage{graphicx}
\usepackage{amsmath}
\usepackage{amssymb}
\usepackage{booktabs}
\usepackage{multirow}
\usepackage{bm}
\usepackage{adjustbox}
\usepackage{pifont}
\usepackage{diagbox}
\usepackage{xcolor}
\usepackage{amsmath,mathtools}
\usepackage{algorithm}
\usepackage[accsupp]{axessibility}
\usepackage[noend]{algpseudocode}

\algnewcommand{\Break}{\textbf{break}}
\usepackage{mathrsfs}

\usepackage[pagebackref,breaklinks,colorlinks]{hyperref}

\usepackage[capitalize]{cleveref}
\crefname{section}{Sec.}{Secs.}
\Crefname{section}{Section}{Sections}
\Crefname{table}{Table}{Tables}
\crefname{table}{Tab.}{Tabs.}

\AtBeginDocument{
  \abovedisplayskip     =0.5\abovedisplayskip
  \abovedisplayshortskip=0.5\abovedisplayshortskip
  \belowdisplayskip     =0.5\belowdisplayskip
  \belowdisplayshortskip=0.5\belowdisplayshortskip}

\def\cvprPaperID{} 
\def\confName{CVPR}
\def\confYear{2022}

\begin{document}

\title{Detecting Deepfakes with Self-Blended Images}
\author{\stepcounter{footnote}Kaede Shiohara
$\quad$ Toshihiko Yamasaki \\ 
The University of Tokyo\\ 
{\tt\small \{shiohara, yamasaki\}@cvm.t.u-tokyo.ac.jp}
}

\maketitle

\begin{abstract}
    In this paper, we present novel synthetic training data called self-blended images (SBIs) to detect deepfakes. 
    SBIs are generated by blending pseudo source and target images from single pristine images, reproducing common forgery artifacts (\textit{e.g.}, blending boundaries and statistical inconsistencies between source and target images). The key idea behind SBIs is that more general and hardly recognizable fake samples encourage classifiers to learn generic and robust representations without overfitting to manipulation-specific artifacts. We compare our approach with state-of-the-art methods on FF++, CDF, DFD, DFDC, DFDCP, and FFIW datasets by following the standard cross-dataset and cross-manipulation protocols. Extensive experiments show that our method improves the model generalization to unknown manipulations and scenes.
    In particular, on DFDC and DFDCP where existing methods suffer from the domain gap between the training and test sets, our approach outperforms the baseline by 4.90\% and 11.78\% points in the cross-dataset evaluation, respectively.
    Code is available at \url{https://github.com/mapooon/SelfBlendedImages}.
\end{abstract}

\section{Introduction}
\label{sec:intro}

The recent rapid advancement of generative adversarial networks~\cite{goodfellow2014generative,dcgan,wgan,lsgan,sagan,progan,stylegan} (GAN) in computer vision has made it possible to generate realistic facial images.
In particular, techniques called deepfake manipulating the identity, expression, or attributes of a subject are used for entertainment purposes, \textit{e.g.}, smartphone applications or movies; however, they can also be used for malicious purposes, \textit{e.g.}, to create fake news or to falsify evidence. Therefore, the vision community is keenly working on deepfake detection techniques.
 
\begin{figure}[t]
 \centering
 \includegraphics[width=1\linewidth]{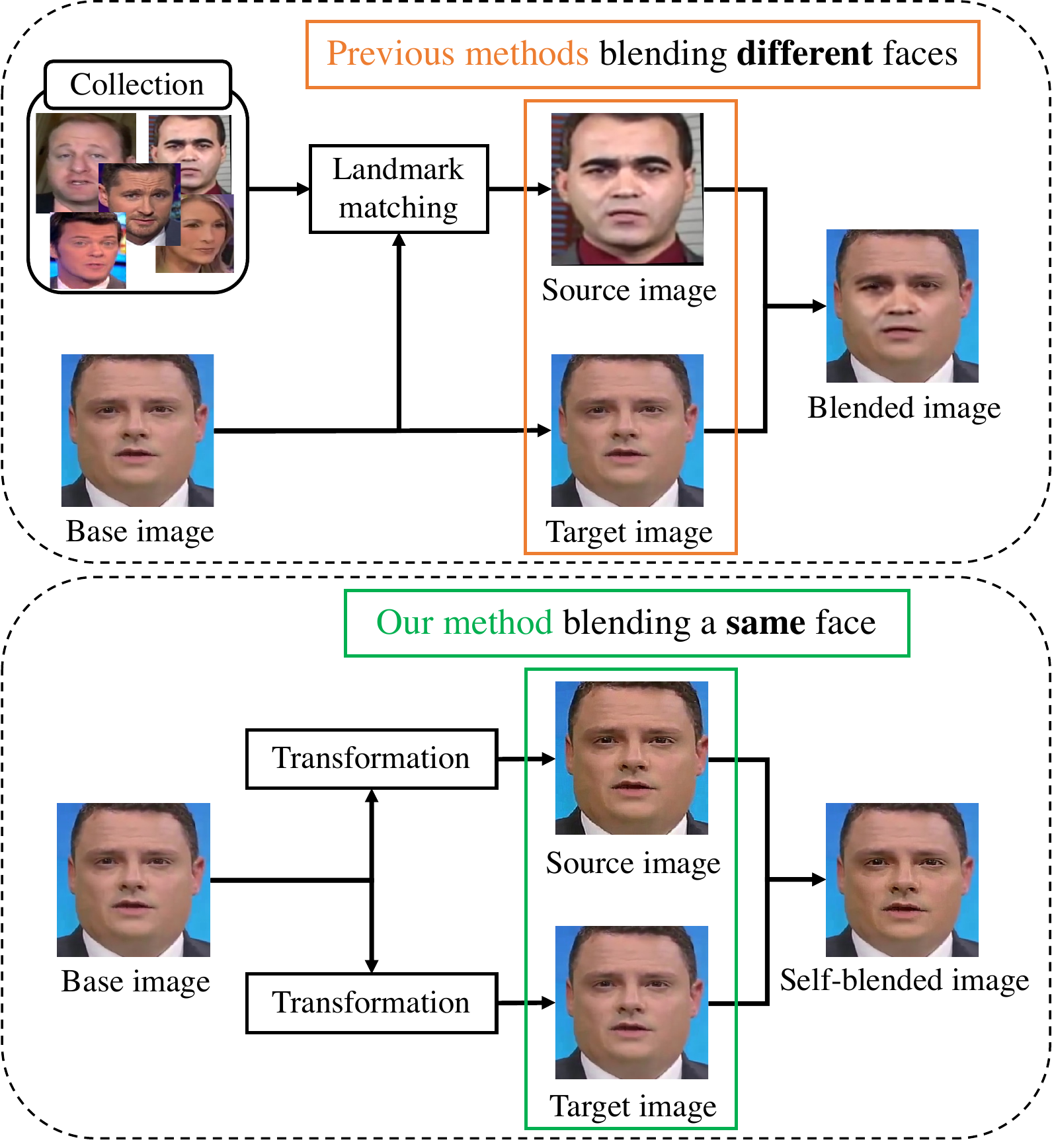}
 \caption{\textbf{Overviews of fake sample syntheses.} Previous methods blend two distinct faces and generate artifacts based on a gap between selected source and target images. By contrast, our method blends slightly changed faces from a single image and generate artifacts actively by transformations. In this example, we apply a color jitter, sharpening, resize, and translation to the source image and no transformations to the target image.}
 \label{fig1}
\end{figure}

Most previous detection methods~\cite{mesonet,dang2020detection,cnnrnn2018,cnnrnn2019,multiattention,blink2020,multitask,kumar2020detecting} perform well on the in-dataset scenario where they detect forgeries they learned in training; however, some studies~\cite{CanTheyBeGeneralized,cozzolino2018forensictransfer,xuan2019generalization,lae} have found that the detection performance significantly drops in the cross-dataset scenario where fake samples are forged by unknown manipulations. 

One of the most effective solutions to this problem is to train models with synthetic data, which encourages models to learn generic representations for deepfake detection.
For example, facial regions are blurred to reproduce a quality degradation of GAN-synthesized source images~\cite{fwa}, blended images are generated from pairs of two pristine images to reproduce blending artifacts~\cite{facexray,pcl}.
However, the quality of deepfakes has improved over the years, which has caused the former method to fail on recent benchmarks~\cite{ffpp,celebdf}. 
Although the latter methods perform well on some datasets~\cite{deepfakedetection,celebdf}, low-quality videos in more challenging datasets~\cite{dfdcp,dfdc} where artifacts are hardly recognizable owing to the high compression or extreme exposure lead them to unacceptable detection performance.

\begin{figure}[t]
    \centering
  \begin{minipage}[b]{0.24\linewidth}
    \centering
    \includegraphics[width=1\linewidth]{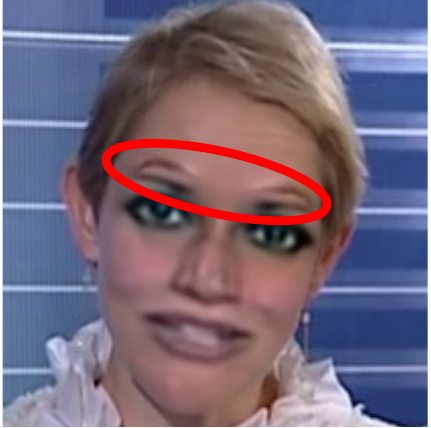}
    \subcaption{Landmark}
  \end{minipage}
  \begin{minipage}[b]{0.24\linewidth}
    \centering
    \includegraphics[width=1\linewidth]{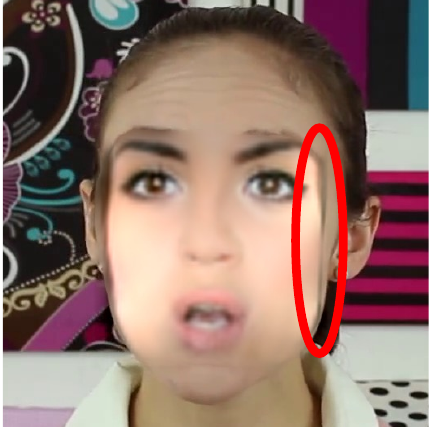}
    \subcaption{Boundary}
  \end{minipage}
  \begin{minipage}[b]{0.24\linewidth}
    \centering
    \includegraphics[width=1\linewidth]{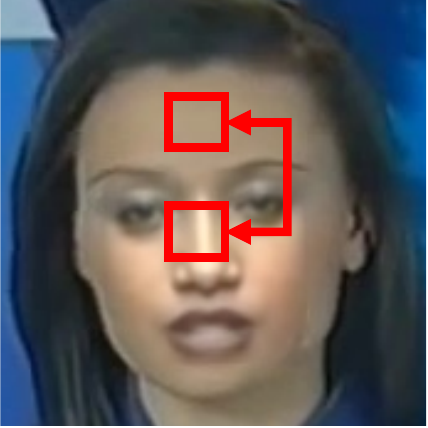}
    \subcaption{Color}
  \end{minipage}
  \begin{minipage}[b]{0.24\linewidth}
    \centering
    \includegraphics[width=1\linewidth]{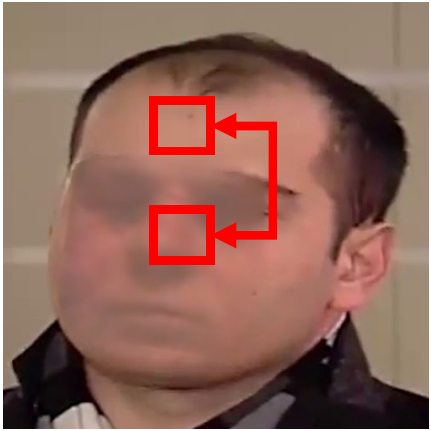}
    \subcaption{Frequency}
  \end{minipage}
  \caption{\textbf{Typical artifacts on forged faces.} We classify artifacts into four types, (a) landmark mismatch, (b) blending boundary, (c) color mismatch, and (d) frequency inconsistency. }
 \label{example_of_artifacts}
\end{figure}

In this paper, we propose novel synthetic training data called self-blended images (SBIs) to detect deepfakes. 
The overviews of our method and previous methods~\cite{facexray,pcl} are shown in Fig.~\ref{fig1}. 
The key idea is that more hardly recognizable fake samples that contain common face forgery traces encourage models to learn more general and robust representations for face forgery detection.
We analyze forged faces and define four typical artifacts motivated from previous works (\textit{e.g.}, blending boundaries~\cite{facexray}, source feature inconsistencies~\cite{pcl}, and statistical anomalies in frequency domain~\cite{chen2021local}) as shown in Fig.~\ref{example_of_artifacts}. 
To synthesize these artifacts based on our key idea, we develop a source-target generator (STG) and mask generator (MG). STG generates pairs of pseudo source and target images from single pristine images using simple image processing, and MG generates various blending masks from facial landmarks of the input images. By blending the source and target images with the masks, we obtain SBIs. 
Training with SBIs encourages the models to learn generic representations because models learn the forgery traces we actively generate in STG. 
Moreover, our method improves training efficiency in terms of computational cost. Whereas successful previous works~\cite{facexray,pcl} use nearest landmark search for source-target pair selection, which is computationally expensive, SBIs are generated without this process. Therefore, our method does not suffer from the large dataset size problem.

We evaluate our approach following the two evaluation protocols,  cross-dataset evaluation and cross-manipulation evaluation. In the cross-dataset evaluation, we train our model on FF++~\cite{ffpp} and evaluate it on CDF~\cite{celebdf}, DFD~\cite{deepfakedetection}, DFDC~\cite{dfdc}, DFDCP~\cite{dfdcp}, and FFIW~\cite{ffiw}. This experimental setting is similar to that in real detection scenarios where defenders are exposed to unseen domains. Our approach surpasses or is at least comparable to the state-of-the-art methods on all test sets despite its simplicity.
Especially, on DFDC and DFDCP where previous methods suffer from domain gaps between the training and test sets, our method outperforms the state-of-the-art unsupervised baseline~\cite{pcl} by 4.90\% and 11.78\% points, respectively.
In the cross-manipulation evaluation, we evaluate the generality of our model on unseen manipulation methods of FF++; DF~\cite{deepfake-faceswap}, F2F~\cite{face2face}, FS~\cite{faceswap}, and NT~\cite{neuraltextures}. Our approach achieves the AUC of 99.99\%, 99.88\%, 99.91\%, and 98.79\% on DF, F2F, FS, and NT, respectively. Although the performance on FF++ becomes saturated, our method still outperforms the state of the art on whole FF++ (99.64\% vs. 99.11\%).

\begin{figure*}[t]
\centering
 \includegraphics[width=0.98\linewidth]
    {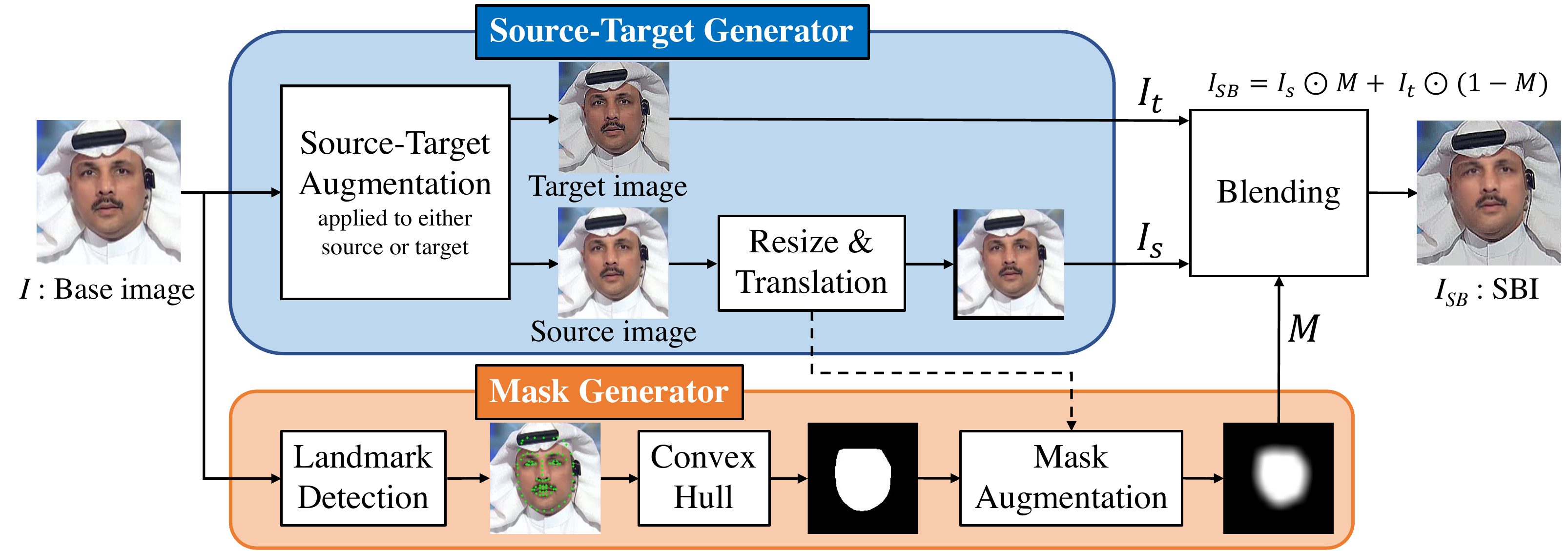}
 \caption{\textbf{Overview of generating an SBI.} A base image $I$ is input into the source-target generator (STG) and mask generator (MG). STG generates pseudo source and target images from the base image using some image transformations, whereas MG generates a blending mask from the facial landmarks and deforms it to increase the diversity of the mask. Finally, the source and target images are blended with the mask.}
 \label{sbi_generation}
\end{figure*}

\section{Related Work}
\noindent \textbf{Deepfake Detection.}
Although many detection methods have been introduced, a primary topic of research is the development of optimal neural network architectures (\textit{e.g.,} an efficient shallow
network\cite{mesonet}, multi-task autoencoders\cite{lae,multitask}, capsule network~\cite{capsule}, recurrent convolutional networks\cite{cnnrnn2018,cnnrnn2019}, and attentional networks\cite{multiattention,dang2020detection}). 
Some studies~\cite{frank2020leveraging,qian2020thinking,li2021frequency,Luo_2021_CVPR,liu2021spatial} focus on the frequency domain to capture forgery traces more effectively. These methods achieve impressive performance on highly compressed videos.
Another notable direction is focusing on specific representations (\textit{e.g.}, head pose~\cite{headpose}, eye blinking \cite{blink2018,blink2020}, mouth movements~\cite{lipfo}, neuron behaviors~\cite{fakespotter}, optical flow~\cite{optflow}, and steganalysis features~\cite{stega}).
Face X-ray~\cite{facexray} introduces a facial representation based on boundaries between altered faces and background images. PCL~\cite{pcl} measures patch-wise similarities of input images to detect inconsistencies between source and target images.
\vskip.5\baselineskip
\noindent \textbf{Training Data Synthesis.}
Although most existing methods perform well in detecting known manipulations, some studies~\cite{CanTheyBeGeneralized,cozzolino2018forensictransfer,xuan2019generalization,lae} have found that the methods cannot be generalized to fake faces forged by unknown manipulations because they tend to overfit to method-specific artifacts seen in training. 
One of the most effective approaches to address this problem is training models with synthetic data; this encourages models to learn generic features for face forgery detection.  
FWA~\cite{fwa} focuses on a quality gap between GAN-synthesized faces and natural faces, and reproduces it on real images by blurring facial regions. However, deepfake techniques have improved over the years and this method fails in detecting forgeries on the recent benchmarks~\cite{ffpp,deepfakedetection}. 
BI~\cite{facexray} and I2G~\cite{pcl} are introduced to generate blended faces which reproduce blending artifacts from pairs of two pristine images with similar facial landmarks.

These blended images work well as fake samples to train more general detection models; however, some concerns remain. 
First, because these blending artifacts depend on  pairs of source and target images selected by the landmark matching, irregular swaps~\cite{examination} are sometimes seen in the generated images. It is possible that these easy samples prevent models from learning robust representations. 
Second, because these methods are introduced to learn the oriented representations \textit{i.e.}, the blending boundary in BI and the source feature consistency in I2G, it is possible that artifacts to be learned for robust deepfake detection are not sufficient only for the artifacts in the blended images.

\section{Self-Blended Images (SBIs)}
Our goal is to detect statistical inconsistencies between altered faces and background images on deepfakes. 
To train more general and robust detectors, we generate synthetic fake samples that consist of common forgery traces, and are difficult to recognize. Our key observation is that, if deepfake generation techniques continue to improve, GAN-synthesized source images will be even closer to pristine target images in their properties, \textit{e.g.}, facial landmarks and pixel statistics. Therefore, we develop a synthetic data generation pipeline where a fake image is generated by blending pseudo source and target images from a single image to give models a more general and difficult task for face forgery detection.

To achieve this, we introduce self-blended images (SBIs). As shown in Fig.~\ref{sbi_generation}, an SBI is generated by three steps; (1) A \textbf{source-target generator} generates pseudo source and target images for blending. The source and target images are augmented to generate statistical inconsistencies (\textit{e.g.}, color and frequency) between them. The source image is also resized and translated to reproduce blending boundaries and landmark mismatches. (2) A \textbf{mask generator} generates a gray-scale mask image with some deformations. (3) We \textbf{blend} the source and target images with the mask to obtain an SBI.
Although the general flow of an SBI generation is illustrated in Fig.~\ref{sbi_generation}, 
we show the pseudocode in Alg.~\ref{algo:pseudo_code} where the procedure is 
slightly different from that in Fig.~\ref{sbi_generation} for training efficiency (\textit{e.g.}, facial landmarks are extracted in preprocess but not in training).
Our pipeline to generate an fake sample has a constant running time regardless of the dataset size while previous methods~\cite{facexray,pcl} have a running time of $O(NK)$ in the preprocess due to the pair selection for source and target images, where $N$ and $K$ are the number of videos and the number of frames of each, respectively~\footnote{Because the official source code of~\cite{facexray,pcl} is not publicly available, we only discuss qualitatively.}.
\begin{algorithm}[t!]
    \caption{Pseudocode for SBIs Generation}
    \textbf{Input}: Base image $I$ of size $(H,W,3)$, facial landmarks $L$ of size $(81,2)$\\
    \textbf{Output}: Self-blended image $I_{SB}$ of size $(H,W,3)$
    \begin{algorithmic}[1]
        \Procedure{$\mathcal{T}$}{$I$} \textbf{:} 
        \Comment{Source-Target Augmentation}
        \State $I \leftarrow \text{ColorTransform}(I)$
        \State  $I \leftarrow \text{FrequencyTransform}(I)$
        \State\Return $I$
        \EndProcedure
        \If {$\text{Uniform}(\text{min}=0,\text{max}=1)<0.5$}
        \State  $I_s, I_t \leftarrow \mathcal{T}(I), I$
        \Else \textbf{ :}
        \State  $I_s, I_t \leftarrow I, \mathcal{T}(I)$
        \EndIf
        \State $I_s, p \leftarrow \text{RandomResizeTranslate}(I_s)$
        \Comment{$p:$ Parameter}
        \State $L \leftarrow \text{LandmarkTransform}(L)$
        \State $M \leftarrow \text{ConvexHull}(L)$
        \State $M \leftarrow \text{ParameterizedResizeTranslate}(M,p)$
        \State $M \leftarrow \text{MaskDeform}(M)$
        \State $r \leftarrow \text{Uniform}(\{0.25,0.5,0.75,1,1,1\})$
        \State $M \leftarrow rM$
        \State $I_{SB} \leftarrow I_s \odot M + I_t \odot (1-M)$
    \end{algorithmic}
    \label{algo:pseudo_code}
\end{algorithm}

\subsection{Source-Target Generator (STG)}
Given an input image $I$, STG initializes pseudo source and target images by copying $I$. To generate statistical inconsistencies between source and target images, STG randomly applies some image transformations to either of them. 
Here, we randomly shift the  values of RGB channels, hue, saturation, value, brightness, and contrast of input images as color transformations. Then we downsample or sharpen input images as frequency transformations. 

To reproduce blending boundaries and landmark mismatches, STG \textbf{resizes} the source image. 
Let the height and width of $I$ be $H$ and $W$, respectively. 
We define the height $H_r$ and width $W_r$ of the resized image as $H_r = u_h H$ and $W_r = u_w W$, 
where $u_h$ and $u_w$ are sampled independently from a continuous uniform distribution $U_{[u_{\text{min}},\ u_{\text{max}}]}$ in the range $[u_{\text{min}},\ u_{\text{max}}]$. The resized image is zero-padded or center-cropped to have the same size as the original.
Then, STG \textbf{translates} the resized source image. We define a translation vector $\bm{t} = [t_h, t_w]$ as $t_h = v_h H$ and $t_w = v_w W$, where $v_h$ and $v_w$  are sampled independently from $U_{[v_{\text{min}},\ v_{\text{max}}]}$.

\subsection{Mask Generator (MG)}
MG provides a gray scale mask image to blend source and target images. To perform this, MG applies a landmark detector to the input image to predict a facial region and initializes a mask by calculating convex hull from predicted facial landmarks. Then the mask is deformed with the landmark transformation as used in BI~\cite{facexray}.
To increase the diversity of blending masks, the shape of masks and blending ratio are randomly changed. First, the mask is deformed by elastic deformation as adopted in~\cite{pcl}. Second, the mask is smoothed by two Gaussian filters with different parameters. After the first smoothing, the pixel values less than 1 are changed to 0. This means that the mask is eroded if the kernel size of the first Gaussian filter is larger than that of the second one and is dilated in the opposite case.
Finally, MG varies the blending ratio of the source image. This can be achieved by multiplying the mask image by a constant $r \in (0, 1]$. Here, we uniformly sample $r$ from $\{0.25, 0.5, 0.75, 1, 1, 1\}$.

\begin{figure}[t]
 \centering
 \includegraphics[width=1.0\linewidth]{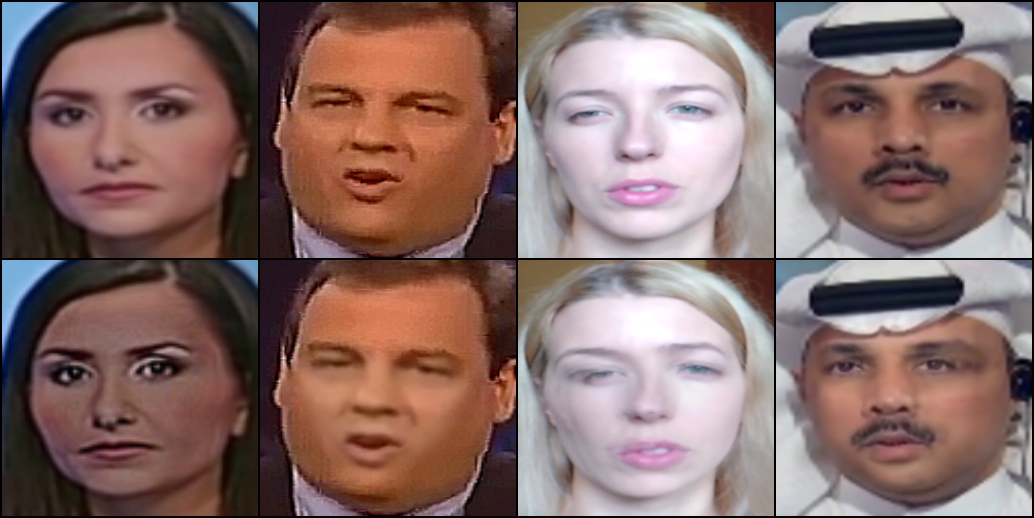}
 \caption{\textbf{Samples of pristine images (top row) and their SBIs (bottom row).}}
 \label{example_of_sbi}
\end{figure}

\subsection{Blending}
By blending the source image $I_s$ and the target image $I_t$  with the blending mask $M$, we obtain the self-blended image $I_{SB}$ as
\begin{equation}
\label{blending}
 I_{SB} = I_{s}\odot M + I_{t}\odot (1-M).
\end{equation}
We show some representative examples of SBIs in Fig.~\ref{example_of_sbi}. Although the purpose of SBIs is not for counterfeiting, they contain artifacts seen in forged faces.

\subsection{Training with SBIs}
Once SBIs are generated, we can train any binary classifier, regardless of whether it is designed for deepfake detection or not. Given input images $X=[x_0,x_1,\cdots,x_{N-1}]$ of size $(N,H,W,3)$ and the corresponding binary labels $T=[t_0,t_1,\cdots,t_{N-1}]$ of size $N$, a classifier $F$ is optimized on the binary cross-entropy loss $L$ as follows:
\begin{equation}
  L \!=\! - \frac{1}{N}\! \sum_{i=0}^{N-1} \{~\!t_i \log F(x_i) \!+\! (1\!-\!t_i) \log (1\!-\!F(x_i))\},
\end{equation}
where $F(x)$ is the probability of $x$ being ``fake''. We input target images as ``Real'' instead of using the base images to encourage the models to focus only on artifacts on SBIs.
Because MG provides blending masks,  we can also adopt mask-based multi-task learning~\cite{facexray,pcl,multitask}.

\begin{table*}[t]
    \small
    \centering
    \begin{adjustbox}{width=0.8\textwidth}
        \begin{tabular}{lcccccccc} \toprule
        \multirow{2}{*}{Method} & \multirow{2}{*}{Input Type}&\multicolumn{2}{c}{Training Set} & \multicolumn{5}{c}{Test Set AUC (\%)}\\ 
        \cmidrule(lr){3-4}\cmidrule(lr){5-9}
        &&Real&Fake& CDF &DFD& DFDC & DFDCP &FFIW\\
        \midrule
        DSP-FWA~\cite{fwa}& Frame&\checkmark &\checkmark&69.30&- & -& - &- \\
        Face X-ray + BI~\cite{facexray} &Frame&\checkmark& &- &93.47&-&71.15 &- \\
        Face X-ray + BI~\cite{facexray} &Frame&\checkmark&\checkmark &- &95.40&-&\underline{80.92} &- \\
        LRL~\cite{chen2021local} &Frame&\checkmark&\checkmark&78.26&89.24  &-&76.53& -  \\
        FRDM~\cite{Luo_2021_CVPR} &Frame&\checkmark&\checkmark &79.4&91.9&-&79.7& -  \\
        
        PCL + I2G~\cite{pcl} &Frame&\checkmark&&\underline{90.03}&\textbf{99.07}&67.52 &74.37& - \\
        \midrule
        Two-branch~\cite{masi2020two}& Video&\checkmark &\checkmark&76.65&- & - &- & -  \\
        DAM~\cite{ffiw} &Video &\checkmark&\checkmark&75.3 &-&- &72.8& -  \\
        LipForensics~\cite{lipfo} & Video&\checkmark &\checkmark&82.4 &-&- &- & -  \\
        FTCN ~\cite{temptrans}& Video&\checkmark &\checkmark&86.9 &94.40$^*$&\underline{71.00}$^*$ &  74.0 &\underline{74.47}$^*$\\
        \midrule
        EFNB4 + SBIs (Ours) &Frame&\checkmark&&\textbf{93.18} & \underline{97.56} &\textbf{72.42} &\textbf{86.15}&\textbf{84.83} \\
        
        \bottomrule
        \end{tabular}
    \end{adjustbox}
  \caption{\textbf{Cross-dataset evaluation on CDF, DFD, DFDC, DFDCP, and FFIW.} The results of prior methods are directly cited from the original paper and their subsequences for fair comparison. Bold and underlined values correspond to the best and the second-best value, respectively. * denotes our experiments with the official code. Our method outperforms state-of-the-art methods on CDF, DFDC, DFDCP, and FFIW, and achieves the second best on DFD without any special network architecture for deepfake detection.}
  \label{tb:cross-dataset}
\end{table*}

\section{Experiments}

\subsection{Implementation Details} 
\label{implementation_details}
\noindent \textbf{Preprocess.} We adopt Dlib~\cite{dlib} and RetinaFace~\cite{retinaface} to extract facial landmarks and bounding boxes from each video frame, respectively. We use an 81 facial landmarks shape predictor\cite{81_landmarks} in Dlib. For the width and height of the face calculated from the bounding box, the face region is cropped with a random margin of 4--20\% for training and a fixed value of 12.5\% for inference.
Note that the landmarks are not needed during inference; hence we only use RetinaFace at the inference time.
\vskip.5\baselineskip
\noindent \textbf{Source-Target Augmentation.}
For the color and frequency transformations, we adopt RGBShift, HueSaturationValue, RandomBrightnessContrast, Downscale, and Sharpen from a widely used image processing toolbox\cite{alb}.
\vskip.5\baselineskip
\noindent \textbf{Training.} We adopt the state-of-the-art convolutional network architecture EfficientNet-b4~\cite{efficientnet} (EFNB4) pre-trained on ImageNet~\cite{imagenet} as the classifier and train it for 100 epochs with the SAM~\cite{sam} optimizer. The batch size and learning rate are set to 32 and 0.001, respectively. 
We sample only eight frames per video for training. If two or more faces are detected in a frame, the one with the largest bounding box is extracted.
Each batch consists of real images and their SBIs, and the same augmentation is applied to each real image and its SBI. 
We also use some data augmentations, \textit{i.e.}, ImageCompression, RGBShift, HueSaturationValue, and RandomBrightnessContrast.
\vskip.5\baselineskip
\noindent \textbf{Model Validation.} Considering practical situations, it is important to validate the model without additional evaluation datasets. We use a validation set that consists of real videos and their SBIs after each epoch, and select the weight with the highest number of epochs among the five weights with the highest AUC. Therefore, no manipulated images are used even for the model validation in our approach.
\vskip.5\baselineskip
\noindent \textbf{Inference Strategy.} We sample 32 frames per video for inference. If two or more faces are detected in a frame, the classifier is applied to all faces and the highest fakeness confidence is used as the predicted confidence for the frame. Once the predictions for all frames are obtained, we average them to get the prediction for the video. For a fair comparison, we use all videos of all test sets for evaluation by setting the confidences to 0.5 for the videos where no face is detected in all frames.

\vskip.5\baselineskip
\subsection{Experimental Setting}
\label{experimental_setting}
\noindent \textbf{Datasets.} We adopt the widely used benchmark \textbf{FaceForensics++}~\cite{ffpp} (FF++) for training, following the convention. It contains 1,000 original videos and 4,000 fake videos forged by four manipulation methods, \textit{i.e.}, Deepfakes~\cite{deepfake-faceswap} (DF), Face2Face~\cite{face2face} (F2F), FaceSwap~\cite{faceswap} (FS), and NeuralTextures~\cite{neuraltextures} (NT). 
For our cross-dataset evaluation, we use five recent deepfake datasets. \textbf{Celeb-DF-v2}~\cite{celebdf} (CDF) applies a more advanced deepfake technique to celebrity videos downloaded from YouTube. \textbf{DeepFakeDetection}~\cite{deepfakedetection} (DFD) provides thousands of deepfake videos generated with consenting actors.
\textbf{DeepFake Detection Challenge Preview}~\cite{dfdcp} (DFDCP) and \textbf{DeepFake Detection Challenge} public test set~\cite{dfdc} (DFDC), that are released along with the competition~\cite{dfdc_compe}, contain a lot of disturbed videos, \textit{e.g.}, compression, downsampling, and noise.
We further provide a novel cross-dataset baseline on a more recent large scale benchmark \textbf{FFIW-10K}~\cite{ffiw} (FFIW) which focuses on multi-person scenario. 
We follow the official train/test splits for all datasets except FFIW where we use the original validation set as our test set because the official test set has not been released yet. 
Although FaceShifter~\cite{faceshifter} and DeeperForensics-1.0~\cite{deeperforensics} provide sophisticated deepfake videos, we do not adopt them in our cross-dataset evaluation because they generate deepfakes from the real videos of FF++ that is the same domain as used in training. 
See the supplementary material for more statistical details.
\vskip.5\baselineskip
\noindent \textbf{Frame-Level Baselines.} We refer to five state-of-the-art frame-level detection methods, including: (1) \textbf{DSP-FWA}~\cite{fwa} propose a training data generation method based on the degradation of the GAN-synthesized source image quality. (2) \textbf{Face X-ray}~\cite{facexray} detects deepfakes via segmenting blending boundaries between source and target images. The model is trained with synthetic fake samples called \textbf{BI} generated by blending two images from different videos. (3) \textbf{Local relation learning}~\cite{chen2021local} (LRL) and (4) \textbf{Fusion + RSA + DCMA + Multi-scale}~\cite{Luo_2021_CVPR} (FRDM) fuse two different representations from RGB and frequency domains. 
(5) \textbf{Pair-wise self-consistency learning}~\cite{pcl} (PCL) detects deepfakes via measuring consistencies between patches of input images. The model is trained with \textbf{inconsistency image generator} (I2G) that is similar to BI~\cite{facexray}.
\vskip.5\baselineskip
\noindent \textbf{Video-Level Baselines.} We further compare our method with video-level methods that output a single scalar fakeness score for some video frames. Unlike frame-level methods, video-level methods can detect incoherence across frames, although they require multiple frames of the subject at regular intervals. We refer to four state-of-the-art methods, including: (1) \textbf{Two-branch}~\cite{masi2020two} proposes Laplacian of Gaussian kernels to enhance the frequency component of the input image. 
(2) \textbf{Discriminative attention model}~\cite{ffiw} (DAM) proposes an attention~\cite{vaswani2017attention}-based network for multi-person scenarios. (3) \textbf{LipForensics}~\cite{lipfo} detects temporal inconsistencies of mouth movements using a pretrained lip-reading model~\cite{martinez2020lipreading}.
(4) \textbf{Fully temporal convolution network}~\cite{temptrans} (FTCN) enhances temporal representations by reducing the spatial kernel size of the convolutions to 1.
\vskip.5\baselineskip
\noindent \textbf{Evaluation Metrics.}
We report the video-level area under the receiver operating characteristic curve (AUC) to compare with prior works. Typically, frame-level predictions are averaged over video frames. We additionally provide average precision (AP)
in the supplementary material.

\begin{table}[t]
    \centering
    \begin{adjustbox}{width=0.475\textwidth}
    \begin{tabular}{lcccc|c} \toprule
      \multirow{2}{*}{Method} & \multicolumn{5}{c}{Test Set AUC (\%)}\\ 
      \cmidrule(lr){2-6}
      &DF & F2F & FS & NT & FF++\\
      \midrule
      Face X-ray + BI~\cite{facexray} & 99.17 & 98.57 & 98.21 & 98.13 & 98.52\\
      PCL + I2G~\cite{pcl} &\textbf{100}& 98.97& 99.86 & 97.63 & 99.11\\
      \midrule
      EFNB4 + SBIs (Ours) &99.99 & \textbf{99.88} & \textbf{99.91} & \textbf{98.79} & \textbf{99.64}\\
      \bottomrule
    \end{tabular}
    \end{adjustbox}
  \caption{\textbf{Cross-manipulation evaluation on FF++.} Our method achieves state-of-the-art results on F2F, FS, NT, and whole FF++. }
  \label{tb:cross-manipulation}
\end{table}
\subsection{Cross-Dataset Evaluation}
To show the generality of our method, we conduct a cross-dataset evaluation where models are trained on FF++ and evaluated on other datasets. 
Although many researchers have considered this task, the test sets used by each of them in their experiments vary from work to work, making comprehensive comparisons difficult. We, therefore, examine the experimental settings in previous works carefully and compile them into Table~\ref{tb:cross-dataset}.
\vskip.5\baselineskip
\noindent \textbf{Comparison with Frame-Level Methods.} Here, we compare our method with other frame-level methods~\cite{fwa,facexray,chen2021local,Luo_2021_CVPR,pcl}. 
Our approach outperforms the state-of-the-art methods on CDF, DFDC, and DFDCP by 6.08\%, 5.17\%, and 5.23\% points, respectively, and improves the baseline by 4.58\% points on average (87.33\% vs. 82.75\%).
Our result is comparable with PCL + I2G~\cite{pcl} on DFD (97.56\% vs. 99.07\%), where a forged face is sometimes placed with some other pristine faces in a manipulated frame, and the percentage of frames that subject is throughout manipulated videos is smaller than other test sets. Therefore, our method can be improved by incorporating any object tracking process into our inference strategy as in PCL + I2G~\cite{pcl}, instead of extracting frames from the video at equal intervals as in our simple strategy.

\vskip.5\baselineskip
\noindent \textbf{Comparison with Video-Level Methods.} We then compare our method with video-level methods~\cite{masi2020two,ffiw,lipfo,temptrans}. 
For more comprehensive comparison, we conduct additional experiments for FTCN~\cite{temptrans} on unconsidered test sets, \textit{i.e.}, DFD, DFDC, and FFIW with officially released code~\cite{ftcn_repo}. The results are denoted as * in Table~\ref{tb:cross-dataset}. 
Our method still outperforms the state of the art by 6.28\%, 3.16\%, 1.42\%, 12.15\%, and 10.36\% points on CDF, DFD, DFDC, DFDCP, and FFIW, respectively, and improves the baseline by 6.68\% points on average (86.83\% vs. 80.15\%). We also evaluate our method on a subset of DFDC which is used in an experiment for LipForensics\cite{lipfo}, outperforming the competitor (76.78\% vs. 73.5\%). The video list is available at the author's repository~\cite{lipfo_repo}.


\subsection{Cross-Manipulation Evaluation}
In real detection situations, the defenders generally are not aware of  the attacker's forgery methods. For this reason, it is important to verify the model generalization to various forgery methods. Following the evaluation protocol used in~\cite{facexray,pcl}, we evaluate our model on four manipulation methods of FF++, \textit{i.e.}, DF, F2F, FS, and NT. We use the raw version for evaluation as well as the competitors.

Table~\ref{tb:cross-manipulation} presents our cross-manipulation evaluation result on FF++. Our method outperforms or nearly equals the existing methods on four manipulations (99.99\% on DF, 99.88\% on F2F, 99.91\% on FS, and 98.79\% on NT) 
and achieves the best performance on the whole FF++ (99.64\% vs. 99.11\%). This result shows that our method works well not only on deepfakes but also on other face manipulations.

\begin{table}[t]
    \centering
    \begin{adjustbox}{width=0.475\textwidth}
    \begin{tabular}{lcccc|c} \toprule
      \multirow{2}{*}{Method} & \multicolumn{5}{c}{Test Set AUC (\%)}\\ 
      \cmidrule(lr){2-6}
      &DF & F2F & FS &NT  & FF++\\
      \midrule
      
      Xception + BI~\cite{facexray}& 98.95 & 97.86 & 89.29 & 97.29 & 95.85\\
      \midrule
     
      Xception + SBIs (Ours)& \textbf{99.99} & \textbf{99.90} & \textbf{98.79} & \textbf{98.20} & \textbf{99.22}\\ 
      \bottomrule
    \end{tabular}
    \end{adjustbox}
    \caption{\textbf{AUC comparison with BI~\cite{facexray}}. 
    }
    \label{tb:compare_with_bi}
\end{table}

\begin{table}[t]
    
    \centering
    \begin{adjustbox}{width=0.475\textwidth}
    \begin{tabular}{lccc|c} \toprule
      \multirow{2}{*}{Method} & \multicolumn{4}{c}{Test Set AUC (\%)}\\
      \cmidrule(lr){2-5}
      &CDF & DFDC & DFDCP & Avg \\
      \midrule
      
      ResNet-34 + I2G~\cite{pcl}& 78.18& 51.72 &69.93 &  66.61\\
      \midrule
  
    ResNet-34 + SBIs (Ours)& \textbf{87.04} &  \textbf{66.41} &\textbf{82.16}&\textbf{78.54}\\ 
      \bottomrule
    \end{tabular}
    \end{adjustbox}
    \caption{\textbf{AUC comparison with I2G~\cite{pcl}}. 
    }
    \label{tb:compare_with_i2g}
\end{table}

\subsection{Data Quality Assessment}
\label{assessment}
Here, we compare our method with state-of-the-art synthetic training data~\cite{facexray,pcl}, removing influences of the difference of the classifiers. To achieve this, we train the same models and optimizer as the ones competitors use in their original papers. 
Table~\ref{tb:compare_with_bi} presents the comparison with BI~\cite{facexray}. 
We train Xception~\cite{xception} with the Adam~\cite{adam} optimizer.
Our method outperforms BI~\cite{facexray} on all the manipulation methods of FF++ in terms of AUC. In particular, the baseline on FS is improved from 89.29\% to 98.79\%. 
Next, the result of the comparison with I2G~\cite{pcl} is given in Table~\ref{tb:compare_with_i2g}. We train ResNet-34~\cite{resnet} with the Adam optimizer. Our method outperforms I2G~\cite{pcl} on CDF, DFDC, and DFDCP and their average by 8.86\%, 14.69\%, 12.23\%, and 11.93\% points, respectively. These results clearly show that our method is superior to the competitors 
as synthetic training data, regardless of the network architecture.

\begin{table}[t]
    \centering
    \begin{adjustbox}{width=0.475\textwidth}
    \begin{tabular}{lcccc|c} \toprule
      \multirow{2}{*}{Process} & \multicolumn{5}{c}{Test Set AUC (\%)}\\ 
      \cmidrule(lr){2-6}
      &FF++ & CDF & DFDCP & FFIW &Avg\\
      \midrule
      w/o Source aug. & 98.58 & \textbf{93.59} &78.06 &61.11 &82.84\\
      w/o Target aug. & 99.35 & 76.61 & 83.84 & 82.87 &\underline{85.67}\\
      w/o S-T aug. & 89.18 & 70.68 & \underline{85.16} & \textbf{88.31} &83.33\\
      w/o Res.\&Trans. & \underline{99.58} & 85.28 & 81.04 & 74.69 &85.15\\
      \midrule
      SBIs (Ours) & \textbf{99.64} & \underline{93.18} & \textbf{86.15} & \underline{84.83} &\textbf{90.95}\\
      \bottomrule
    \end{tabular}
    \end{adjustbox}
  \caption{\textbf{Effect of each process of STG.} The skipping of any process causes a fatal performance degradation.}
  \label{tb:process_ablation}
\end{table}

\begin{table}[t]
    \centering
    \begin{adjustbox}{width=0.475\textwidth}
    \begin{tabular}{cccccc|c} \toprule
        \multicolumn{2}{c}{Training Set} & \multicolumn{5}{c}{Test Set AUC (\%)}\\ 
       \cmidrule(lr){1-2}\cmidrule(lr){3-7}
        Database & \#Real&FF++ &CDF& DFDCP& FFIW &Avg \\
        \midrule
        FF++ &720&{\color{brown}99.64}&93.18&86.15&84.83&90.95 \\
        CDF &622 &98.10 &{\color{brown}93.74}&81.10&77.82&87.69\\
        DFDCP&737 &98.76 &90.79&{\color{brown}88.70} &81.31&89.89\\
        FFIW& 7090&99.72&95.57&78.91&{\color{brown}88.07} &90.57\\
      \bottomrule
    \end{tabular}
    \end{adjustbox}
    \caption{\textbf{Performance of different training datasets.} 
    Our method achieves good results on each training dataset. ``\#Real'' presents the number of real videos of the training set, excluding that of the validation set.}
  \label{tb:dataset_ablation}
\end{table}

\subsection{Ablations}
\label{ablations}
\vskip.5\baselineskip
\noindent \textbf{Effect of Each Process of STG.} In STG, we use some image processing to generate pseudo source and target images. Conversely, because learned representations are based on the artifacts we actively provide in STG, ablation experiments of the generation process enable the exploration of effective clues on the deepfake benchmarks. 
Here, we train our model without some processes, \textit{i.e.}, the source augmentation, target augmentation, source-target augmentation, or resize and translation, and evaluate them on FF++, CDF, DFDCP, and FFIW.
As shown in Table~\ref{tb:process_ablation}, source and target augmentation is indeed effective in detecting deepfakes, and both of them are necessary for better performance. We also observe that the resize and translation reproduce important artifacts because of the poor performance without them.
Through the ablation, it can be concluded that different clues are useful to detectors on different datasets because they have different deepfake generation processes.

\vskip.5\baselineskip
\noindent \textbf{Generality to Training Datasets.} It is important from a practical standpoint to show that our method can perform well on various real face datasets. We here train models with SBIs from the pristine videos of FF++, CDF, DFDCP, and FFIW. Then we evaluate them on the test sets. On CDF and FFIW, we split the original training sets into the alternative training/validation sets.
Table~\ref{tb:dataset_ablation} presents the result. Our method is generalized to all datasets without a critical performance drop. We observe the large dataset size of FFIW contributes to the model generality. However, the difference of video scene between FFIW and DFDCP leads to a slight performance drop on DFDCP; FFIW consists of videos collected from YouTube, whereas DFDCP consists of videos made by filming recruited subjects.
The result also indicates that learning pristine videos can help detect forged faces in the same domain as that in training, even if models did not learn manipulated videos, as indicated by the scores highlighted in brown in Table~\ref{tb:dataset_ablation}, which supports our not adopting FaceShifter~\cite{faceshifter} and DeeperForensics-1.0~\cite{deeperforensics} in the cross-dataset evaluation, as mentioned in Section~\ref{experimental_setting}.

\begin{table}[t]
    \begin{adjustbox}{width=0.475\textwidth}
    \begin{tabular}{lcccc|c} \toprule
        \multirow{2}{*}{Architecture} & \multicolumn{5}{c}{Test Set AUC (\%)}\\ 
      \cmidrule(lr){2-6}
      &FF++ &CDF& DFDCP & FFIW&Avg\\
      \midrule
       ResNet-50 & 97.77& 90.66 &82.88&79.30& 87.65\\
       ResNet-152 &98.33 &90.71&\underline{85.01}& 76.43& 87.62\\
        Xception & \underline{99.26}&90.27&78.85&76.72 &86.28\\
       EfficientNet-b1& 99.10&\underline{91.16}&84.58&\underline{80.23} &\underline{88.77}\\
       EfficientNet-b4& \textbf{99.64}&\textbf{93.18}&\textbf{86.15}&\textbf{84.83} &\textbf{90.95}\\
      \bottomrule
    \end{tabular}
    \end{adjustbox}
    \caption{\textbf{Performance of different network architectures.} An architecture with larger capacity tends to result in better generality.}
  \label{tb:model_ablation}
\end{table}

\vskip.5\baselineskip
\noindent \textbf{Choice of Network Architecture.} Although we adopt EfficientNet-b4\cite{efficientnet} as our standard classifier, our method can be applied to other network architectures. Here, we investigate the performance of different state-of-the-art architectures, \textit{i.e.}, ResNet-50, -152~\cite{resnet}, Xception~\cite{xception}, EfficientNet-b1, and -b4\cite{efficientnet} trained with SBIs. As shown in Table~\ref{tb:model_ablation}, all architectures achieve good results on FF++, CDF, DFDCP, and FFIW without critical performance degradation. Remarkably, even our method with a vanilla ResNet-50 outperforms all the previous methods on CDF, DFDCP, and FFIW as shown in Tables~\ref{tb:cross-dataset} and ~\ref{tb:model_ablation}.
We observe larger networks tend to result in greater generality, which indicates SBIs provide a variety of training samples.

\begin{figure*}[t]
\centering
 \includegraphics[width=1.0\linewidth]
      {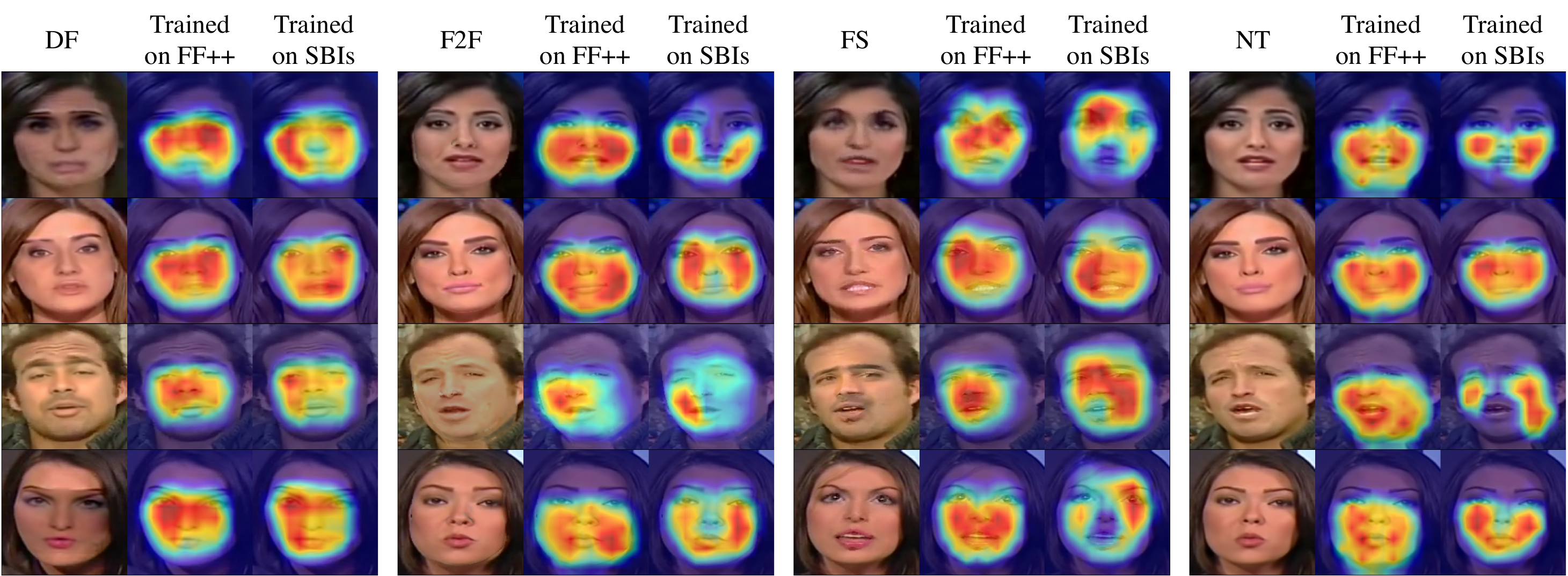}
 \caption{\textbf{Saliency map visualization of the baseline and our model.} The baseline captures method-specific artifacts that are widely present across forged faces while our model detects minor artifacts independent of manipulations. Best viewed in color.}
 \label{gradcam}
\end{figure*}

\begin{figure}[t]
\centering
 \includegraphics[width=0.99\linewidth]
      {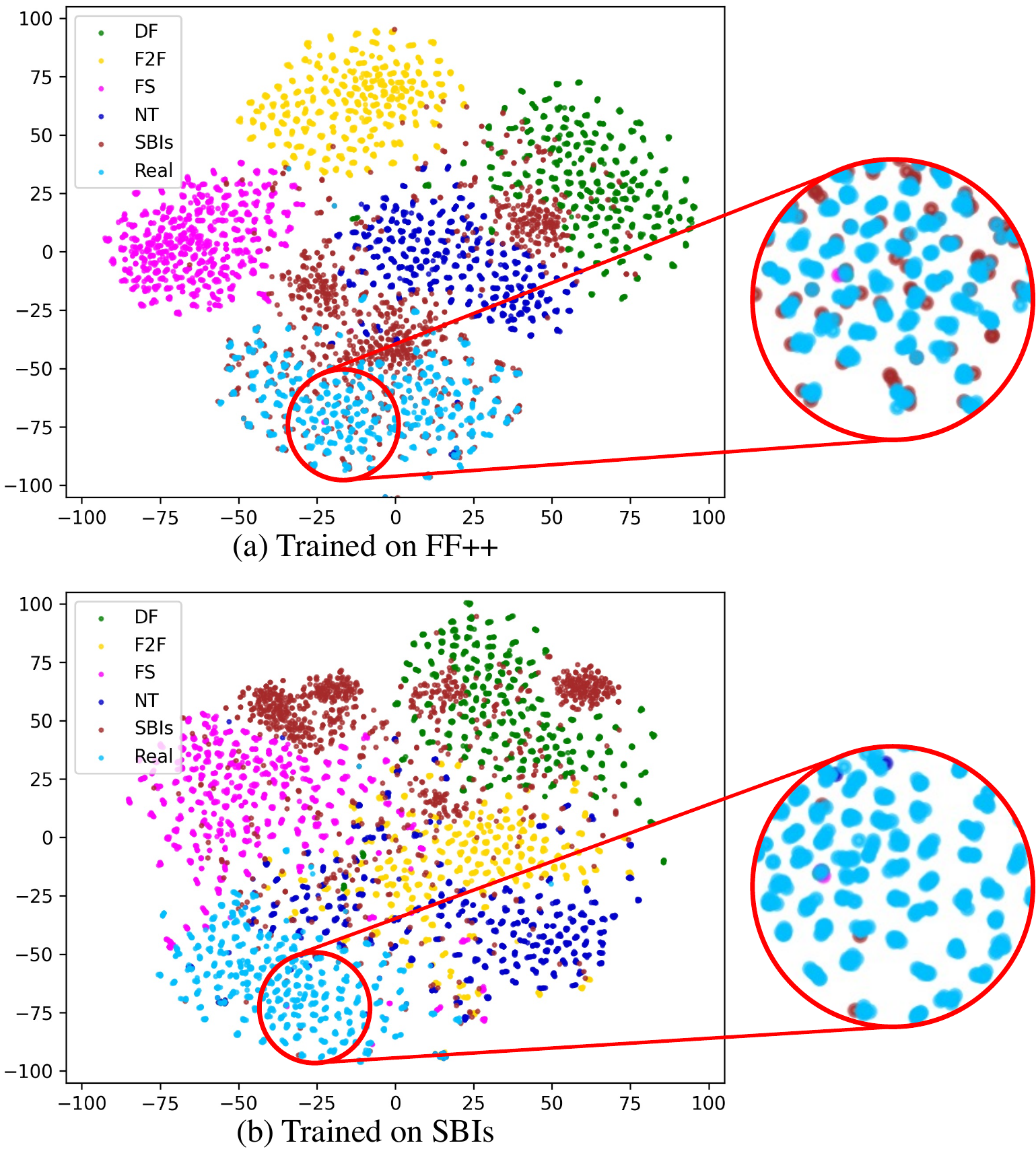}
 \caption{\textbf{Feature space visualization of the baseline (a) and our model (b).} The baseline cannot distinguish real images from SBIs (because the feature vectors fall into the same feature space) while our model succeeds in distinguishing real images from not only SBIs but also forged images. Best viewed in color.}
 \label{tsne}
\end{figure}

\subsection{Qualitative Analysis}
To obtain qualitative insights, we visualize model saliency maps and feature spaces. Through the analysis, we use two models; one is trained on FF++ (baseline) and the other is trained on SBIs (our model).

\vskip.5\baselineskip
\noindent \textbf{Saliency Map.} To visualize where the models are paying their attention on the forged faces, we apply GradCAM++~\cite{gradcampp} to the models on manipulated frames of FF++, \textit{i.e.}, DF, F2F, FS, and NT, as shown in Fig.~\ref{gradcam}. It can be observed that our method encourages the model to make its attentions sparser than the baseline. This is because our model detects minor artifacts independent of manipulations, e.g., blending boundaries, while the baseline captures method-specific pixel distributions that are widely spread in the forged faces.

\vskip.5\baselineskip
\noindent \textbf{Feature Space.} We then apply t-SNE~\cite{tsne} visualization to feature vectors from the last layers of the models. We emphasize again that it is easy for the baseline to recognize the forged faces because they are seen in its training, and that our goal is to separate real faces from others, not to classify types of manipulations. As shown in Fig.~\ref{tsne}, the baseline cannot distinguish SBIs from real images although it clusters four manipulations seen in training. On the other hand, our model distinguishes not only SBIs but also forged faces from real ones. We also observe that SBIs are distributed all over the four manipulations in the feature space. These results indicate that SBIs are general synthetic data to train face forgery detectors.

\section{Limitations}
Although our results in cross-dataset and cross-manipulation evaluations are expected to be beneficial, we observe some limitations of our method. 
First, similar to other frame-level methods, our model cannot capture temporal inconsistencies across video frames. Therefore, sophisticated deepfake generation techniques with fewer spatial artifacts may pass our detector. 
Moreover, our method does not perform well on whole-image synthesis because we define a ``fake image'' as an image where the face region or background is manipulated. We evaluate our model on a 20k image set sampled from FFHQ dataset and StyleGAN~\cite{stylegan} synthesis, and its AUC is only 69.11\%.

\section{Conclusion}

In this paper, we proposed a novel synthetic training data, self-blended images (SBIs), based on the idea that more general and hardly recognizable fake samples encourage classifiers to learn more generic and robust representations. SBIs are generated by blending pseudo source and target images that are slightly transformed from single real images to reproduce forgery artifacts. 
Using SBIs, we could train detectors without forged face images.
Extensive experiments show that our method is superior to state-of-the-art methods for unseen manipulations and scenes, and generalized to different network architectures and training datasets.

\appendix

\section{Additional Experiments}
To show the validity and generality of our proposed method, we conduct some additional experiments using BI~\cite{facexray}. We implement it following the author's code~\cite{bi_repo}.
\vskip.5\baselineskip
\noindent \textbf{Landmark Similarity.}  In BI, 100 images with the closest facial landmarks to the base image are used as candidate source images. We train EfficientNet-B4~\cite{efficientnet} (EFNB4) on all 100 (BI$_{\text{1-100}}$) and on the top 20 (BI$_{\text{1-20}}$). 
As shown in Table~\ref{tb:additional}(a), the model trained on BI$_{\text{1-20}}$ outperforms the model trained on original BI on CDF and DFDCP, and is on par with on DFD, DFDC, and FFIW. This result indicates that, at least in the landmark similarity, easy samples with low similarity do not contribute to the model generality.

\vskip.5\baselineskip
\noindent \textbf{Joint Training of SBIs and BI.}
To explore the best practice for more general deepfake detection, we train EFNB4 on the joint dataset of SBIs and BI that are each sampled with the probability of 0.5. The results of joint-training are basically lower than that of our proposed SBIs as shown in Table~\ref{tb:additional}(b). 

\vskip.5\baselineskip
\noindent \textbf{Applying Source-Target Augmentation to BI.} To demonstrate the superiority of our two ideas of (1) blending identical images and (2) augmenting source and target images, we incorporate our source-target augmentation into BI. As shown in Table~\ref{tb:additional}(c), the results are lower than that of SBIs on four out of the five test sets, although they are better than that of the original BI, which indicates our two ideas are both important for general deepfake detection.

\section{Comprehensive Results}
We provide comprehensive results of our method including video-level area under the receiver operating characteristic curve (AUC) and average precision (AP). We also describe the number of real and fake videos. We additionally evaluate our method on FaceShifter~\cite{faceshifter} (FSh) and DeeperForensics1.0~\cite{deeperforensics} (DF1.0) datasets. On DF1.0, we use c23 (lightly compressed) real videos of FF++ following the convention. The result is given in Table~\ref{tb:comprehensive_result}.

\newpage

\begin{table}[H]
    \begin{adjustbox}{width=0.475\textwidth}
    \begin{tabular}{lccccc} \toprule
        \multirow{2}{*}{Method} & \multicolumn{5}{c}{Test Set AUC (\%)}\\ 
      \cmidrule(lr){2-6}
      &CDF&DFD &DFDC& DFDCP&FFIW \\
      \midrule
      \multicolumn{6}{l}{(a) Effect of Landmark Similarity}\\
      \midrule
       BI$_{\text{1-100}}$ (original) &69.40&\textbf{97.50}&\textbf{66.55}&68.71&\textbf{85.69}\\
      BI$_{\text{1-20}}$&\textbf{71.44}  &  97.27 &65.27& \textbf{68.80}  &85.35 \\
      \midrule
      \midrule
        \multicolumn{6}{l}{(b) Effect of Joint-Training of SBIs and BI}\\
       \midrule
      SBIs (Ours) &\textbf{93.18} & 97.56 &\textbf{72.42} &\textbf{86.15}&84.83 \\
       BI &69.40&97.50&66.55&68.71&\textbf{85.69}\\
       SBIs + BI&89.36&\textbf{98.34}&71.87&82.92&83.53\\
       \midrule
       \midrule
       \multicolumn{6}{l}{(c) Effect of Source-Target Augmentation in BI}\\
       \midrule
       SBIs (Ours) &\textbf{93.18} & 97.56 &\textbf{72.42} &\textbf{86.15}&84.83 \\
       BI &69.40&97.50&66.55&68.71&\textbf{85.69} \\
       BI w/ S-T Aug.&78.99&\textbf{99.04}&71.98&74.25&83.14\\
      \bottomrule
    \end{tabular}
    \end{adjustbox}
    \caption{\textbf{Additional experiments.}}
  \label{tb:additional}
\end{table}

\begin{table}[H]
    \centering
    \begin{adjustbox}{width=0.475\textwidth}
    
        \begin{tabular}{lcccc} \toprule
        \multicolumn{3}{c}{Test Set}\rule{0pt}{2.3ex}&\multicolumn{2}{c}{Metrics}\\ 
        \cmidrule(lr){1-3}\cmidrule(lr){4-5}
        Database&\#Real&\#Fake&AUC(\%)&AP(\%)\rule{0pt}{2.3ex}\\
        \midrule
        DF~\cite{deepfake-faceswap} &140    &140    &99.99  &99.99  \rule{0pt}{2.3ex}\\
        F2F~\cite{face2face}    &140    &140    &99.88  &99.89\rule{0pt}{2.3ex}\\
        FS~\cite{faceswap}  &140    &140    &99.91  &99.91\rule{0pt}{2.3ex}\\
        NT~\cite{neuraltextures}    &140    &140    &98.79  &99.15\rule{0pt}{2.3ex}\\
        FF++~\cite{ffpp}    &140    &560    &99.64  &99.92  \rule{0pt}{2.3ex}\\
        DFD~\cite{deepfakedetection}    &363    &3068   &97.56  &99.70  \rule{0pt}{2.3ex}\\
        FSh~\cite{faceshifter}  &140    &140    &98.27  &98.24  \rule{0pt}{2.3ex}\\
        DF1.0~\cite{deeperforensics}    &140    &140    &83.14  &85.06  \rule{0pt}{2.3ex}\\
        CDF~\cite{celebdf}  &178    &340    &93.18  &96.35  \rule{0pt}{2.3ex}\\
        DFDC~\cite{dfdc}    &2500   &2500   &72.42  &75.17  \rule{0pt}{2.3ex}\\
        DFDCP~\cite{dfdcp}  &276    &501    &86.15  &91.37  \rule{0pt}{2.3ex}\\
        FFIW~\cite{ffiw}    &250    &250    &84.83  &84.30  \rule{0pt}{2.3ex}\\
        \bottomrule
        \end{tabular}
    \end{adjustbox}
  \caption{\textbf{Comprehensive results and statistical details.} }
  \label{tb:comprehensive_result}
\end{table}

{\clearpage
\small
\bibliographystyle{ieee_fullname}
\bibliography{egbib}
}

\end{document}